\documentclass[letterpaper]{article} 
\usepackage[preprint]{aaai2027}  
\usepackage[hyphens]{url}  
\usepackage{graphicx} 
\usepackage{natbib}  
\usepackage{caption} 
\usepackage{algorithm}
\usepackage{algorithmic}

\usepackage{newfloat}
\usepackage{listings}
\DeclareCaptionStyle{ruled}{labelfont=normalfont,labelsep=colon,strut=off} 
\floatstyle{ruled}
\newfloat{listing}{tb}{lst}{}
\floatname{listing}{Listing}

\usepackage{booktabs}
\usepackage{multirow}

\title{SSR: Similarity-Shift Refinement for Training-Free Object-Centric Masks}
\author{
    Xiaoqian Lu\textsuperscript{\rm 1},
    Guangfu Guo\textsuperscript{\rm 2}
}

\affiliations{
    \textsuperscript{\rm 1}University of Ottawa
    \quad
    \textsuperscript{\rm 2}Clemson University\\
}

\begin{document}

\maketitle

\begin{abstract}
Object-centric models often produce fragmented masks, boundary leakage, and incorrect region merging. We introduce Similarity-Shift Refinement (SSR), a training-free post-hoc method for improving object-centric masks with a frozen self-supervised Vision Transformer. SSR measures changes in pairwise patch similarity before and after self-attention value aggregation, retains positively strengthened relations, and constructs a sparse affinity graph. This graph propagates the initial soft slot assignments in a single refinement step, without retraining or modifying either model. Across natural-image, synthetic-video, and real-world-video benchmarks, SSR improves all-pixel Adjusted Rand Index in all 24 evaluated model–dataset combinations, with an average gain of 8.5 percentage points. Ablations show that value-space similarity shifts outperform query- and key-space variants as well as static Transformer affinities. However, texture-dense scenes may cause visually similar regions to be over-grouped. Overall, SSR provides a simple and transferable signal for training-free object-centric mask refinement.

\end{abstract}


\section{Introduction}

Object-centric learning models decompose visual scenes into a set of
object-level slot representations and corresponding masks
\citep{locatello2020slot}. These models have progressed from synthetic
scenes to more challenging natural images by reconstructing self-supervised
visual features or introducing stronger slot-learning objectives
\citep{seitzer2023bridging,kakogeorgiou2024spot}. However, the resulting
slot masks often remain noisy in real-world scenes. A single object may be
fragmented across multiple slots, adjacent objects may be incorrectly
merged, and foreground regions may leak into visually similar background
areas. These errors reduce the usefulness of object-centric representations
for unsupervised segmentation and downstream scene-understanding tasks.

A straightforward solution is to retrain the object-centric model or add a
model-specific mask-refinement module. Such approaches, however, commonly
depend on the architecture, training objective, and checkpoint of the base
model. In parallel, self-supervised Vision Transformers have been shown to
encode object-related structures in their patch features and attention maps
\citep{caron2021dino,oquab2023dinov2}. Training-free object-discovery
methods such as LOST and TokenCut use frozen patch representations or
attention-derived affinities for object localization and graph-based
segmentation \citep{simeoni2021lost,wang2022tokencut}. Nevertheless, these
methods primarily rely on static information already encoded in a feature
space or attention map. Unlike static affinity methods that ask which patches are already similar, SSR measures which patch relations are strengthened by a patch-restricted attention aggregation derived from the frozen Transformer.

Our key observation is that self-attention does not merely produce another
set of patch representations; it changes the relationships among patches
during value aggregation. Unlike static affinity methods that ask
which patches are already similar, SSR asks which patches become more
similar after attention aggregation. We treat such positive relational
changes as contextual grouping signals for refining the initial slot
assignments. Based on this observation, we propose SSR, a
training-free post-hoc method for refining object-centric masks. SSR
extracts positive value-space relational shifts from selected layers and
heads of a frozen DINOv2 encoder, converts them into a sparse patch-level
propagation graph, and uses this graph to update the initial soft slot
assignments. It does not modify the object-centric model, the DINOv2
encoder, or either model checkpoint.

Experiments across multiple image and video object-centric baselines show
that SSR provides a transferable refinement signal. It consistently
improves all-pixel clustering quality and produces particularly stable
improvements on natural-image and real-world-video benchmarks, where noisy
slot masks frequently contain fragmented regions and background leakage.
However, the signal is not uniformly reliable across visual domains. In
texture-dense scenes such as ClevrTex, positive relational shifts may follow
repeated texture or background co-occurrence rather than object-instance
boundaries. Unconstrained propagation can consequently over-group visually
similar but semantically distinct regions, improving all-pixel consistency
while degrading foreground and overlap metrics. We treat this behavior as
an important applicability boundary of the proposed signal.

Our contributions are summarized as follows:
\begin{itemize}
    \item We introduce attention-induced relational shift, which captures
    patch relations strengthened by value aggregation in a frozen
    self-supervised Transformer.

    \item We propose SSR, a training-free method that propagates existing
    soft slot assignments over the resulting relation graph.

    \item We validate SSR across image and video models, distinguish it
    from simpler attention and affinity baselines, and analyze its
    texture-induced failure mode on ClevrTex.
\end{itemize}

\section{Related Work}

\paragraph{Object-Centric Learning.}
Slot Attention represents a scene with an unordered set of object-like
slots \citep{locatello2020slot}. Later methods improve object discovery
through stronger decoders, self-supervised feature reconstruction, and
more stable slot optimization, including SLATE, DINOSAUR, SlotDiffusion,
SPOT, DIAS, and SmoothSA
\citep{singh2021slate,seitzer2023bridging,wu2023slotdiffusion,
kakogeorgiou2024spot,zhao2025dias,zhao2025smoothsa}. Video methods such
as VideoSAUR, SlotContrast, and RandSF.Q additionally model temporal
consistency \citep{zadaianchuk2023videosaur,manasyan2024slotcontrast,
zhao2025randsfq}. Unlike these training-based approaches, SSR refines the
soft masks of an already trained model without changing its architecture,
objective, or checkpoint.

\paragraph{Self-Supervised Transformers for Object Discovery.}
Self-supervised Vision Transformers encode object-related structure in
their patch features and attention maps
\citep{caron2021dino,oquab2023dinov2}. Training-free methods such as LOST
and TokenCut use frozen patch similarities or attention-derived graphs for
object localization and segmentation
\citep{simeoni2021lost,wang2022tokencut}. These approaches mainly rely on
static information already present in a feature or attention space. SSR
instead measures how pairwise patch similarity changes after attention
aggregates the value tokens, using the strengthened relations to refine an
existing multi-slot decomposition.

\paragraph{Training-Free Propagation and Mask Refinement.}
Attention rollout, attention flow, and Transformer attribution propagate
attention or relevance to estimate token influence
\citep{abnar2020attentionflow,chefer2021transformer}. Classical refinement
methods instead use RGB consistency or static feature graphs, as in
DenseCRF, label propagation, and normalized cuts
\citep{krahenbuhl2011densecrf,zhou2004localglobal,shi2000normalizedcuts}.
SSR neither propagates raw attention weights nor directly uses static
feature similarity. It constructs a graph from positive value-space
relational shifts and propagates existing slot assignments over this
graph.

\section{Method}
\label{sec:method}

\begin{figure*}[t]
    \centering
    \includegraphics[width=0.98\textwidth]{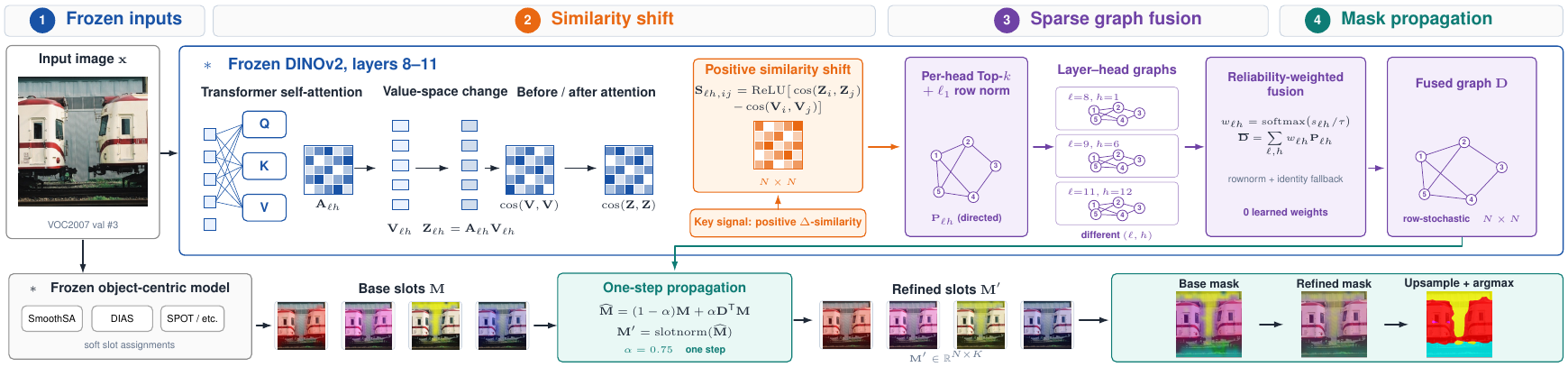}
\caption{
\textbf{Overview of SSR.}
SSR extracts positive attention-induced changes in patch similarity from
a frozen DINOv2 encoder, builds a sparse fused graph, and uses one-step
propagation to refine the soft masks of a frozen object-centric model.
}
\label{fig:ssr_overview}
\end{figure*}

\begin{algorithm}[t]
\caption{SSR Mask Refinement}
\label{alg:delta_v}
\footnotesize
\begin{algorithmic}[1]
\STATE \textbf{Input:} Soft slot assignments $\mathbf{M}$,
frozen DINOv2 blocks, layers $\mathcal{L}$

\FOR{each layer $\ell\in\mathcal{L}$ and head $h$}
    \STATE Extract patch values $\mathbf{V}_{\ell h}$
    and attention $\mathbf{A}_{\ell h}$

    \STATE Compute attended values:
    $\mathbf{Z}_{\ell h}
    =\mathbf{A}_{\ell h}\mathbf{V}_{\ell h}$

    \STATE Compute similarity shifts:
    $\mathbf{\Delta}_{\ell h}
    ={\rm ReLU}\bigl(
    \cos(\mathbf{Z}_{\ell h})
    -\cos(\mathbf{V}_{\ell h})
    \bigr)$

    \STATE Keep the row-wise Top-$k$ positive values:
    $\widetilde{\mathbf{\Delta}}_{\ell h}
    ={\rm TopK}^{+}_{k}(\mathbf{\Delta}_{\ell h})$

    \STATE Normalize the graph:
    $\mathbf{P}_{\ell h}
    =\widetilde{\mathbf{\Delta}}_{\ell h}
    \big/
    \sum_j\widetilde{\Delta}_{\ell h,ij}$

    \STATE Compute the reliability score $s_{\ell h}$
\ENDFOR

\STATE Compute fusion weights:
$w_{\ell h}
=
\frac{e^{s_{\ell h}/\tau}}
{\sum_{\ell',h'}e^{s_{\ell'h'}/\tau}}$

\STATE Fuse the graphs:
$\mathbf{D}
=
\sum_{\ell,h}w_{\ell h}\mathbf{P}_{\ell h}$

\STATE Propagate slot assignments:
$\widehat{\mathbf{M}}
=
(1-\alpha)\mathbf{M}
+\alpha\mathbf{D}^{T}\mathbf{M}$

\STATE Normalize across slots:
$M'_{ik}
=
\widehat{M}_{ik}
\big/
\sum_{k'}\widehat{M}_{ik'}$

\STATE \textbf{Return:}
Refined masks using $\arg\max_k M'_{ik}$
\end{algorithmic}
\end{algorithm}

\paragraph{Problem Setting.}

Consider a frozen object-centric model that produces soft slot assignments
$\mathbf{M}\in\mathbf{R}^{N\times K}$, where each row is a probability
distribution over $K$ slots, i.e., $M_{ik}\geq0$ and
$\sum_{k=1}^{K}M_{ik}=1$.

Our goal is to refine $\mathbf{M}$ without updating the object-centric model
or its checkpoint. If the native mask resolution differs from the DINOv2
patch grid, we bilinearly resize the soft assignments to this grid and
renormalize them across slots.

We process the same input with a frozen DINOv2 ViT-S/14 encoder with
register tokens \citep{oquab2023dinov2,darcet2024registers}. For layer
$\ell$ and attention head $h$, let
$\mathbf{V}_{\ell h}\in\mathbf{R}^{N\times d_h}$ denote the value tokens
after head splitting. We remove the class and register tokens before
constructing the patch-only attention matrix. Using the
normalized and scaled patch queries and keys, we reconstruct the
patch-only, post-softmax attention matrix
$\mathbf{A}_{\ell h}\in\mathbf{R}^{N\times N}$.
Each row of $\mathbf{A}_{\ell h}$ therefore sums to one over the patch
tokens. DINOv2 is used in evaluation mode, so attention dropout is
disabled. The attended-value representation is
{\small
\begin{equation}
\mathbf{Z}_{\ell h}
=
\mathbf{A}_{\ell h}\mathbf{V}_{\ell h}.
\label{eq:attended_value}
\end{equation}
}
Here, $\mathbf{Z}_{\ell h}$ is computed before the output projection,
residual connection, and MLP.

\paragraph{Attention-Induced Relational Shift.}

For each patch pair $(i,j)$, we compare their cosine similarities before and
after attended-value aggregation:
{\small
\begin{equation}
\Delta_{\ell h,ij}
=
{\rm ReLU}
\left(
\cos(\mathbf{Z}_{\ell h,i},\mathbf{Z}_{\ell h,j})
-
\cos(\mathbf{V}_{\ell h,i},\mathbf{V}_{\ell h,j})
\right).
\label{eq:delta_v}
\end{equation}
}
The vectors are $\ell_2$-normalized, and cosine similarity is computed
independently within each attention head. A positive value indicates that
the two patches become more similar after value aggregation. We discard
negative changes because SSR propagates assignments only along
relations strengthened by attention.

\paragraph{Sparse Graph Construction and Fusion.}

For each row of $\mathbf{\Delta}_{\ell h}$, we first set the diagonal to
zero and retain at most the $k$ largest strictly positive entries. If fewer
than $k$ positive entries exist, all available positive entries are kept.
Let $\widetilde{\mathbf{\Delta}}_{\ell h}$ denote the resulting sparse
graph. We normalize it by its row sum:
{\small
\begin{equation}
\begin{array}{rcl}
\widetilde{\mathbf{\Delta}}_{\ell h}
& = &
\mathrm{TopK}^{+}_{k}(\mathbf{\Delta}_{\ell h}),
\\[3pt]
P_{\ell h,ij}
& = &
\displaystyle
\frac{\widetilde{\Delta}_{\ell h,ij}}
{\max\left(
\sum_{j'}\widetilde{\Delta}_{\ell h,ij'},\epsilon
\right)}.
\end{array}
\label{eq:sparse_affinity}
\end{equation}
}
This is $\ell_1$ row normalization rather than an edge-wise softmax, and no
temperature is applied to individual graph edges.

Different heads are weighted according to the concentration and magnitude
of their positive relational shifts. For each layer-head graph, we first normalize the raw positive shifts:
{\small
\begin{equation}
p_{\ell h,ij}
=
\frac{
\Delta_{\ell h,ij}
}{
\max
\left(
\sum_{j'}\Delta_{\ell h,ij'},
\epsilon
\right)
}.
\label{eq:relation_probability}
\end{equation}
}
We define the dispersion and average magnitude of the positive shifts as
{\small
\begin{equation}
\begin{array}{rcl}
\overline{H}_{\ell h}
& = &
-\displaystyle\frac{1}{N\log N}
\sum_{i,j}
p_{\ell h,ij}
\log\left(p_{\ell h,ij}+\epsilon\right),
\\[4pt]
\overline{m}_{\ell h}
& = &
\displaystyle\frac{1}{N^2}
\sum_{i,j}
\Delta_{\ell h,ij}.
\end{array}
\label{eq:relation_statistics}
\end{equation}
}
The corresponding reliability score is
{\small
\begin{equation}
s_{\ell h}
=
\left(
1-\overline{H}_{\ell h}
\right)
\log
\left(
1+\overline{m}_{\ell h}
\right).
\label{eq:reliability_score}
\end{equation}
}
This score favors positive relational shifts that are sufficiently strong
without being overly dispersed. We compute
{\small
\begin{equation}
w_{\ell h}
=
\frac{
\exp(s_{\ell h}/\tau)
}{
\sum_{\ell',h'}
\exp(s_{\ell'h'}/\tau)
},
\qquad
\overline{\mathbf{D}}
=
\sum_{\ell,h}w_{\ell h}\mathbf{P}_{\ell h}.
\label{eq:fusion_weights}
\end{equation}
}
Here, $\tau$ is used only for the layer-head weight softmax. We
row-normalize $\overline{\mathbf{D}}$ to obtain $\mathbf{D}$. If a fused row
is empty, it is replaced by the corresponding identity row.

The graph is intentionally not symmetrized. Although
$\mathbf{\Delta}_{\ell h}$ is symmetric before sparsification, row-wise
Top-$k$ selection produces a directed graph in which each patch selects its
own strongest neighbors. The final matrix $\mathbf{D}$ is nonnegative and
row-stochastic.

\paragraph{Slot Assignment Propagation.}

Because $D_{ij}$ represents propagation from source patch $i$ to target
patch $j$, SSR updates the assignments as
{\small
\begin{equation}
\widehat{\mathbf{M}}
=
(1-\alpha)\mathbf{M}
+
\alpha\mathbf{D}^{T}\mathbf{M},
\qquad
M'_{ik}
=
\frac{
\widehat{M}_{ik}
}{
\max
\left(
\sum_{k'}\widehat{M}_{ik'},
\epsilon
\right)
}.
\label{eq:propagation}
\end{equation}
}
The refined soft assignments $\mathbf{M}'$ are bilinearly upsampled to the
evaluation resolution before taking the slot-wise $\arg\max$. All metrics
are computed at this resolution. We do not apply nearest-neighbor
upsampling to hard patch labels or any connected-component post-processing.

\paragraph{Semantic/local-boundary SSR.}

Let $\mathbf{C}$ denote the nonnegative cosine similarity of
frozen DINOv2 tokens averaged over the last four blocks.
Let $\mathbf{S}_b$ retain the row-wise Top-$k_b$ entries of
$\mathbf{C}$. For $b\in\{h,s\}$, we define
{\small
\begin{equation}
\begin{array}{rcl}
\mathbf{G}_b
& = &
f_b\mathbf{1}
+(1-f_b)\mathbf{C}\odot\mathbf{S}_b,
\\[3pt]
\mathbf{D}^{b}
& = &
\mathrm{RowNorm}
\bigl(\mathbf{D}\odot\mathbf{G}_b\bigr),
\end{array}
\label{eq:semantic_branches}
\end{equation}
}
where $(k_h,f_h)=(48,0.05)$ and
$(k_s,f_s)=(96,0.30)$.

Let $m$ denote the average mass of $\mathbf{D}^{s}$ within
the Top-$48$ token-similarity neighbors. We adapt the graph
and propagation strength as
{\small
\begin{equation}
\begin{array}{rcl}
r
& = &
\mathrm{clip}
\bigl((m-0.30)/0.15,\,0,\,1\bigr),
\\[3pt]
\mathbf{D}^{\mathrm{sem}}
& = &
\mathrm{RowNorm}
\bigl((1-r)\mathbf{D}^{h}+r\mathbf{D}^{s}\bigr),
\\[3pt]
\alpha
& = &
0.50+0.25r.
\end{array}
\label{eq:semantic_mixing}
\end{equation}
}

Pseudo-superpixels are connected components of adjacent
patches whose RGB--Sobel distance is below the median.
We set $B_{ij}=1$ for patches in the same component and
$B_{ij}=0.30$ otherwise. The final transition is
{\small
\begin{equation}
\begin{array}{rcl}
q
& = &
\mathrm{clip}
\bigl((0.45-r)/0.15,\,0,\,1\bigr),
\\[3pt]
\mathbf{D}^{\mathrm{safe}}
& = &
\mathrm{RowNorm}\bigl(
\mathbf{D}^{\mathrm{sem}}\odot\mathbf{W}_B
\bigr),
\\[3pt]
\mathbf{W}_B
& = &
\mathbf{1}+q(\mathbf{B}-\mathbf{1}).
\end{array}
\label{eq:local_boundary}
\end{equation}
}

The semantic and boundary cues only reweight directed edges.
All parameters are fixed across checkpoints, and no labels are
used at inference.

\section{Experiments}
\label{sec:experiments}

\newcommand{\std}[1]{%
  \textcolor{gray}{(#1)}%
}

\begin{table*}[t]
\centering
\caption{Image object-discovery results on COCO and VOC. Each cell reports
ARI / ARI-FG / mBO / mIoU, with standard deviations shown in gray.
Bold marks the best result within each base-model block and metric.}
\label{tab:main_image_results_compact}

\scriptsize
\setlength{\tabcolsep}{1.6pt}
\renewcommand{\arraystretch}{1.12}

\begin{tabular}{@{}llcccccc@{}}
\toprule
Dataset
& Refiner
& SLATE
& DINOSAUR
& SlotDiff.
& SPOT
& DIAS
& SmoothSA$_i$
\\[-1pt]

\midrule

\multirow{5}{*}{COCO}
& Frozen
& \shortstack{
17.5/28.8/26.8/25.4\\
\std{0.6/0.3/0.3/0.3}}
& \shortstack{
18.2/37.0/28.3/26.9\\
\std{1.0/1.2/0.5/0.5}}
& \shortstack{
17.7/29.0/27.0/25.6\\
\std{0.5/0.1/0.4/0.4}}
& \shortstack{
23.7/40.4/30.9/29.3\\
\std{0.5/0.5/0.2/0.2}}
& \shortstack{
25.6/41.2/31.7/30.2\\
\std{0.1/0.3/0.1/0.1}}
& \shortstack{
29.3/41.3/33.4/31.8\\
\std{1.0/1.2/0.2/0.2}}
\\[1.5pt]

& +DenseCRF
& \shortstack{
23.1/21.6/29.6/28.2\\
\std{0.5/0.2/0.2/0.2}}
& \shortstack{
26.0/25.8/31.4/29.9\\
\std{1.3/0.6/0.4/0.4}}
& \shortstack{
23.1/22.0/29.6/28.1\\
\std{0.6/0.1/0.4/0.3}}
& \shortstack{
28.1/27.1/32.8/31.2\\
\std{0.5/0.2/0.1/0.1}}
& \shortstack{
29.5/27.3/33.1/31.6\\
\std{0.1/0.1/0.1/0.1}}
& \shortstack{
33.8/26.1/33.6/32.0\\
\std{1.2/0.8/0.1/0.2}}
\\[1.5pt]

& +LP
& \shortstack{
22.3/30.7/28.8/27.3\\
\std{0.7/0.1/0.4/0.4}}
& \shortstack{
24.9/39.3/31.1/29.6\\
\std{1.1/1.1/0.4/0.5}}
& \shortstack{
22.4/31.0/28.8/27.3\\
\std{0.5/0.0/0.4/0.3}}
& \shortstack{
25.8/41.6/32.1/30.4\\
\std{0.7/0.4/0.3/0.3}}
& \shortstack{
27.1/41.4/32.6/31.0\\
\std{0.2/0.2/0.1/0.1}}
& \shortstack{
31.1/41.2/33.9/32.3\\
\std{1.1/1.1/0.1/0.2}}
\\[1.5pt]

& +SS-Ncut
& \shortstack{
23.1/30.8/29.2/27.7\\
\std{0.7/0.2/0.4/0.4}}
& \shortstack{
25.8/39.1/31.5/29.9\\
\std{1.2/1.1/0.4/0.5}}
& \shortstack{
23.2/31.1/29.2/27.7\\
\std{0.5/0.1/0.3/0.3}}
& \shortstack{
26.6/41.3/32.3/30.6\\
\std{0.8/0.6/0.3/0.3}}
& \shortstack{
28.1/41.4/33.0/31.4\\
\std{0.2/0.2/0.2/0.2}}
& \shortstack{
32.2/40.9/34.1/32.4\\
\std{1.2/1.1/0.2/0.3}}
\\

\cmidrule(lr){2-8}

& +\textbf{SSR}
& \shortstack{
\textbf{25.1}/\textbf{31.4}/\textbf{30.0}/\textbf{28.6}\\
\std{0.9/0.2/0.4/0.5}}
& \shortstack{
\textbf{27.7}/\textbf{39.6}/\textbf{32.1}/\textbf{30.5}\\
\std{1.4/1.1/0.5/0.5}}
& \shortstack{
\textbf{25.3}/\textbf{31.8}/\textbf{30.1}/\textbf{28.6}\\
\std{0.7/0.2/0.4/0.4}}
& \shortstack{
\textbf{29.1}/\textbf{42.2}/\textbf{33.1}/\textbf{31.3}\\
\std{0.9/0.8/0.3/0.3}}
& \shortstack{
\textbf{31.2}/\textbf{42.8}/\textbf{34.0}/\textbf{32.4}\\
\std{0.3/0.2/0.1/0.1}}
& \shortstack{
\textbf{36.1}/\textbf{42.1}/\textbf{35.0}/\textbf{33.3}\\
\std{1.5/1.3/0.2/0.3}}
\\

\midrule

\multirow{5}{*}{VOC}
& Frozen
& \shortstack{
18.6/26.2/37.2/36.1\\
\std{0.1/0.8/0.5/0.4}}
& \shortstack{
21.5/36.2/40.6/39.7\\
\std{0.7/1.3/0.6/0.6}}
& \shortstack{
17.0/21.7/35.2/34.0\\
\std{1.2/1.8/0.9/1.0}}
& \shortstack{
24.5/31.0/40.1/38.6\\
\std{0.3/0.8/0.2/0.3}}
& \shortstack{
30.9/33.5/43.4/42.4\\
\std{0.5/0.7/0.5/0.5}}
& \shortstack{
35.0/33.6/45.2/43.9\\
\std{0.5/1.2/0.3/0.3}}
\\[1.5pt]

& +DenseCRF
& \shortstack{
23.1/18.3/39.5/38.1\\
\std{0.0/0.8/0.5/0.5}}
& \shortstack{
30.1/23.9/43.7/42.4\\
\std{0.5/0.6/0.4/0.4}}
& \shortstack{
21.8/15.9/38.0/36.5\\
\std{1.0/1.3/0.8/0.8}}
& \shortstack{
29.3/19.2/\textbf{42.8}/41.0\\
\std{0.3/0.9/0.1/0.2}}
& \shortstack{
33.1/21.4/44.3/43.0\\
\std{0.4/0.6/0.2/0.2}}
& \shortstack{
38.3/20.0/45.1/43.5\\
\std{0.5/0.5/0.2/0.2}}
\\[1.5pt]

& +LP
& \shortstack{
23.9/28.2/39.2/37.9\\
\std{0.1/0.7/0.3/0.2}}
& \shortstack{
29.4/37.4/43.8/42.7\\
\std{0.6/1.8/0.4/0.4}}
& \shortstack{
21.9/23.0/37.2/35.8\\
\std{1.3/1.2/1.2/1.2}}
& \shortstack{
26.8/35.8/41.6/40.0\\
\std{0.5/1.3/0.5/0.5}}
& \shortstack{
31.8/33.0/44.1/43.0\\
\std{0.1/0.2/0.3/0.2}}
& \shortstack{
36.3/33.8/45.8/44.4\\
\std{0.2/0.8/0.2/0.2}}
\\[1.5pt]

& +SS-Ncut
& \shortstack{
24.6/27.5/39.8/38.5\\
\std{0.2/1.1/0.2/0.2}}
& \shortstack{
30.7/37.3/44.5/43.4\\
\std{0.6/2.0/0.3/0.3}}
& \shortstack{
22.8/23.1/37.8/36.4\\
\std{1.4/1.3/1.1/1.1}}
& \shortstack{
28.0/35.8/42.1/40.4\\
\std{0.5/1.7/0.5/0.6}}
& \shortstack{
33.2/32.6/44.8/43.7\\
\std{0.2/0.2/0.2/0.1}}
& \shortstack{
37.7/33.5/46.2/44.8\\
\std{0.4/0.6/0.4/0.4}}
\\

\cmidrule(lr){2-8}

& +\textbf{SSR}
& \shortstack{
\textbf{26.9}/\textbf{29.0}/\textbf{41.1}/\textbf{39.9}\\
\std{0.4/0.8/0.0/0.1}}
& \shortstack{
\textbf{33.5}/\textbf{38.3}/\textbf{45.6}/\textbf{44.5}\\
\std{0.6/2.3/0.2/0.2}}
& \shortstack{
\textbf{24.7}/\textbf{24.3}/\textbf{39.0}/\textbf{37.6}\\
\std{1.3/1.0/1.0/1.1}}
& \shortstack{
\textbf{29.6}/\textbf{36.8}/\textbf{42.8}/\textbf{41.2}\\
\std{0.8/1.2/0.4/0.4}}
& \shortstack{
\textbf{36.2}/\textbf{34.9}/\textbf{46.1}/\textbf{45.0}\\
\std{0.4/0.4/0.5/0.4}}
& \shortstack{
\textbf{41.6}/\textbf{35.5}/\textbf{47.5}/\textbf{46.0}\\
\std{0.3/1.2/0.3/0.4}}
\\

\bottomrule
\end{tabular}
\end{table*}

\begin{table*}[t]
\centering
\caption{Video object-discovery results on MOVi-C, MOVi-E, and YTVIS-2022.
Each cell reports ARI / ARI-FG / mBO / mIoU, with standard deviations
shown in gray. Bold marks the best result within each base-model block
and metric.}
\label{tab:main_video_results_compact}

\scriptsize
\setlength{\tabcolsep}{1.6pt}
\renewcommand{\arraystretch}{1.12}

\resizebox{\textwidth}{!}{%
\begin{tabular}{@{}llcccc@{}}
\toprule
Dataset
& Refiner
& VideoSAUR
& SlotContrast
& RandSF.Q
& SmoothSA$_v$
\\
\midrule

\multirow{5}{*}{\rotatebox[origin=c]{90}{MOVi-C}}
& Frozen
& 41.9\std{1.1}/53.3\std{2.1}/16.1\std{0.4}/14.8\std{0.4}
& 64.6\std{9.4}/\textbf{59.9}\std{5.3}/\textbf{27.7}\std{3.0}/\textbf{25.8}\std{2.9}
& 65.4\std{10.7}/\textbf{67.4}\std{2.1}/\textbf{29.2}\std{3.8}/\textbf{26.8}\std{3.7}
& 50.9\std{1.6}/\textbf{69.0}\std{0.3}/31.7\std{0.8}/30.2\std{0.8}
\\[1.5pt]

& +DenseCRF
& \textbf{59.6}\std{9.8}/49.4\std{2.8}/\textbf{21.0}\std{3.3}/\textbf{19.2}\std{3.1}
& 69.9\std{14.8}/53.9\std{7.4}/26.7\std{4.1}/24.6\std{4.0}
& 71.3\std{19.0}/58.3\std{5.6}/28.6\std{6.8}/26.1\std{6.6}
& \textbf{74.4}\std{1.0}/59.4\std{0.9}/\textbf{40.1}\std{0.6}/\textbf{38.1}\std{0.7}
\\[1.5pt]

& +LP
& 47.3\std{5.8}/\textbf{54.5}\std{2.6}/17.7\std{1.8}/16.1\std{1.7}
& 68.6\std{11.4}/59.3\std{6.6}/27.6\std{2.6}/25.6\std{2.5}
& 67.2\std{12.7}/63.7\std{4.0}/28.2\std{3.7}/25.8\std{3.4}
& 56.5\std{1.9}/68.0\std{1.1}/31.5\std{1.0}/29.9\std{1.0}
\\[1.5pt]

& +SS-Ncut
& 50.0\std{7.7}/53.0\std{3.1}/18.0\std{2.2}/16.3\std{2.1}
& 69.3\std{11.9}/58.1\std{6.7}/26.9\std{2.3}/24.8\std{2.1}
& 68.2\std{14.0}/62.1\std{4.5}/27.4\std{3.6}/24.9\std{3.2}
& 59.3\std{2.0}/66.7\std{1.2}/32.0\std{1.0}/30.3\std{1.0}
\\

\cmidrule(lr){2-6}

& +\textbf{SSR}
& 56.4\std{9.6}/53.2\std{3.8}/19.8\std{2.8}/18.0\std{2.6}
& \textbf{70.2}\std{10.9}/56.6\std{9.0}/25.9\std{0.9}/24.0\std{0.9}
& \textbf{71.7}\std{16.8}/63.3\std{5.2}/28.8\std{4.7}/26.3\std{4.4}
& 70.4\std{1.7}/65.9\std{0.4}/36.0\std{0.8}/34.2\std{0.8}
\\

\midrule

\multirow{5}{*}{\rotatebox[origin=c]{90}{MOVi-E}}
& Frozen
& 17.4\std{2.5}/34.6\std{20.7}/8.3\std{4.9}/7.5\std{4.3}
& 29.9\std{4.9}/70.6\std{3.8}/20.7\std{1.4}/19.3\std{1.2}
& 30.5\std{1.2}/\textbf{82.1}\std{3.1}/\textbf{23.0}\std{1.2}/\textbf{21.6}\std{1.4}
& 36.7\std{0.6}/73.6\std{0.6}/28.6\std{0.1}/27.4\std{0.1}
\\[1.5pt]

& +DenseCRF
& 19.9\std{2.8}/34.3\std{21.6}/\textbf{8.6}\std{5.2}/\textbf{7.8}\std{4.6}
& 39.9\std{13.6}/63.6\std{7.2}/21.2\std{3.5}/19.8\std{3.2}
& 29.7\std{0.7}/72.8\std{2.1}/19.2\std{1.0}/17.9\std{1.2}
& \textbf{63.9}\std{0.5}/59.5\std{0.2}/31.6\std{0.1}/30.0\std{0.1}
\\[1.5pt]

& +LP
& 19.2\std{3.4}/\textbf{35.6}\std{21.3}/8.4\std{5.1}/7.6\std{4.4}
& 36.7\std{8.0}/\textbf{70.9}\std{6.6}/22.1\std{1.2}/20.5\std{0.9}
& 31.8\std{1.1}/80.0\std{2.2}/22.6\std{1.1}/21.2\std{1.3}
& 43.8\std{0.8}/\textbf{74.1}\std{0.4}/28.5\std{0.6}/27.1\std{0.6}
\\[1.5pt]

& +SS-Ncut
& 19.8\std{3.8}/34.9\std{20.8}/8.4\std{5.1}/7.5\std{4.4}
& 39.3\std{8.6}/69.6\std{7.2}/22.4\std{1.3}/20.8\std{1.0}
& 32.8\std{0.8}/78.3\std{1.7}/22.6\std{1.1}/21.2\std{1.3}
& 47.0\std{0.8}/72.6\std{0.5}/29.1\std{0.6}/27.7\std{0.6}
\\

\cmidrule(lr){2-6}

& +\textbf{SSR}
& \textbf{20.6}\std{4.5}/35.3\std{21.2}/\textbf{8.6}\std{5.3}/7.7\std{4.6}
& \textbf{42.7}\std{9.1}/70.0\std{7.6}/\textbf{22.5}\std{1.5}/\textbf{21.0}\std{1.3}
& \textbf{33.5}\std{0.5}/81.4\std{2.2}/22.1\std{1.3}/20.7\std{1.4}
& 57.2\std{0.9}/71.9\std{0.5}/\textbf{31.7}\std{0.3}/\textbf{30.3}\std{0.4}
\\

\midrule

\multirow{5}{*}{\rotatebox[origin=c]{90}{YTVIS-2022}}
& Frozen
& 33.4\std{0.8}/48.2\std{0.7}/27.2\std{0.3}/26.8\std{0.3}
& 35.2\std{0.8}/51.4\std{0.7}/29.7\std{0.5}/29.3\std{0.6}
& 37.9\std{1.3}/51.8\std{1.2}/32.2\std{1.8}/31.5\std{1.8}
& 42.0\std{0.6}/59.0\std{2.1}/36.0\std{0.5}/34.9\std{0.6}
\\[1.5pt]

& +DenseCRF
& \textbf{42.9}\std{0.7}/44.3\std{0.7}/32.0\std{0.1}/31.5\std{0.1}
& \textbf{44.4}\std{1.3}/47.3\std{0.3}/34.6\std{0.5}/34.0\std{0.5}
& \textbf{46.6}\std{1.3}/46.0\std{1.4}/35.5\std{1.5}/34.6\std{1.4}
& \textbf{52.2}\std{0.1}/54.4\std{1.9}/38.9\std{0.3}/37.6\std{0.5}
\\[1.5pt]

& +LP
& 37.6\std{0.8}/52.0\std{1.3}/31.0\std{0.3}/30.6\std{0.3}
& 39.3\std{1.1}/55.3\std{0.9}/33.7\std{0.6}/33.1\std{0.6}
& 39.8\std{1.3}/53.4\std{2.0}/33.2\std{1.8}/32.5\std{1.8}
& 44.1\std{0.3}/62.2\std{1.5}/36.9\std{0.4}/35.7\std{0.5}
\\[1.5pt]

& +SS-Ncut
& 39.1\std{0.9}/\textbf{52.4}\std{1.6}/31.9\std{0.3}/31.4\std{0.3}
& 40.7\std{1.2}/\textbf{55.9}\std{0.9}/34.6\std{0.6}/34.0\std{0.6}
& 41.1\std{1.4}/54.3\std{2.4}/34.0\std{2.0}/33.2\std{1.9}
& 45.7\std{0.4}/62.9\std{1.6}/37.5\std{0.5}/36.3\std{0.6}
\\

\cmidrule(lr){2-6}

& +\textbf{SSR}
& 42.1\std{1.2}/\textbf{52.4}\std{1.7}/\textbf{33.4}\std{0.3}/\textbf{32.9}\std{0.4}
& 43.2\std{1.4}/55.8\std{0.6}/\textbf{35.7}\std{0.7}/\textbf{35.2}\std{0.7}
& 44.4\std{1.6}/\textbf{54.8}\std{2.3}/\textbf{35.6}\std{2.0}/\textbf{34.9}\std{1.9}
& 50.4\std{0.4}/\textbf{63.2}\std{1.8}/\textbf{39.6}\std{0.6}/\textbf{38.4}\std{0.7}
\\

\bottomrule
\end{tabular}
}
\end{table*}

\paragraph{Experimental Setup.}

We evaluate SSR as a training-free post-hoc refiner for frozen image and
video object-centric models. COCO is a diverse natural-image benchmark
with complex multi-object scenes \citep{lin2014coco}, while PASCAL VOC
contains natural images with pixel-level object annotations
\citep{everingham2010pascal}. MOVi-C and MOVi-E are synthetic video
benchmarks with increasing scene complexity and object interactions
\citep{greff2022kubric}, whereas YTVIS-2022 evaluates instance
segmentation in real-world videos \citep{yang2019video}. We additionally
use ClevrTex, a texture-rich synthetic benchmark, to study failures caused
by repeated visual patterns \citep{karazija2021clevrtex}. We report Adjusted Rand Index (ARI) for all-pixel clustering and foreground
ARI (ARI-FG) for clustering restricted to foreground pixels
\citep{hubert1985comparing}. Mean Best Overlap (mBO) measures how well each
ground-truth object is covered by its best predicted mask, while mean
Intersection over Union (mIoU) measures average mask overlap after
prediction-to-ground-truth matching. Main results are averaged over three
checkpoint seeds, and each SSR result is paired with the frozen base
checkpoint from the same seed before aggregation.

\paragraph{Implementation Details.}

SSR uses a frozen DINOv2 ViT-S/14 encoder with four register tokens. Inputs are resized to
$224\times224$ and normalized using the standard ImageNet statistics,
producing a $16\times16$ patch grid. We use all attention heads from
zero-indexed layers 8--11 and remove the class and register tokens.
Video frames are processed independently. Unless otherwise specified, SSR uses $k=48$, $\tau=0.1$,
$\alpha=0.75$, and one propagation step. We selected $k=48$, $\tau=0.1$, and $\alpha=0.75$ once
on the PASCAL VOC validation split using three development
checkpoints, and then froze this configuration for all other
models and datasets without dataset-specific retuning.

\paragraph{Base Models and Comparators.}

For images, we evaluate SLATE \citep{singh2021slate}, DINOSAUR
\citep{seitzer2023bridging}, SlotDiffusion \citep{wu2023slotdiffusion},
SPOT \citep{kakogeorgiou2024spot}, DIAS \citep{zhao2025dias}, and
SmoothSA$_i$ \citep{zhao2025smoothsa}. For videos, we evaluate VideoSAUR
\citep{zadaianchuk2023videosaur}, SlotContrast
\citep{manasyan2024slotcontrast}, RandSF.Q \citep{zhao2025randsfq}, and
SmoothSA$_v$ \citep{zhao2025smoothsa}. Together, these baselines cover a
range of image- and video-based object-centric architectures, allowing us
to evaluate whether SSR transfers across different slot-learning
formulations. All base models and checkpoints remain frozen, and no model
is retrained for this work.

We compare SSR with DenseCRF \citep{krahenbuhl2011densecrf},
label propagation (LP) \citep{zhou2004localglobal}, and slot-seeded
normalized cuts (SS-Ncut) \citep{shi2000normalizedcuts}. DenseCRF uses RGB appearance and spatial information,
whereas LP and SS-Ncut construct static affinity graphs using the same
frozen DINOv2 encoder as SSR. SSR differs by constructing its
graph from attention-induced changes in pairwise value-space similarity.
Detailed checkpoint, slot-count, seed, and comparator configurations are
provided in the supplementary material.

\paragraph{Main Results.}

Tables~\ref{tab:main_image_results_compact}
and~\ref{tab:main_video_results_compact} summarize the image and video
results. SSR provides the most consistent gains in all-pixel ARI and
performs particularly well on natural-image and real-world-video
benchmarks. Across the evaluated base-model and dataset pairings, the
all-pixel ARI increases consistently after applying SSR.

On COCO and VOC, SSR improves all four metrics for the evaluated image
models. These results indicate that attention-induced relational shifts can
reconnect fragmented slot regions and reduce isolated background
assignments when the relational graph is aligned with the scene structure.
The gains also transfer across substantially different object-centric
architectures, suggesting that the signal is not tied to one specific slot
decoder or checkpoint.

The video results highlight the consistency of SSR across
models and domains. SSR improves all-pixel ARI in all
$12$ video model--dataset combinations, while also improving
mBO and mIoU in $9$ of $12$ settings. The gains are
particularly balanced on YTVIS-2022, where SSR improves all
four metrics for every evaluated model. Although DenseCRF occasionally achieves a higher ARI, its
behavior is less balanced: large global-clustering gains often
coincide with substantial reductions in ARI-FG. SSR instead
provides a more stable cross-dataset trade-off, consistently
improving global clustering while preserving foreground and
overlap quality more reliably. On the more challenging
MOVi-C and MOVi-E benchmarks, some foreground metrics still
decline, but the degradation is generally less severe than
that produced by aggressive appearance-based refinement.
Overall, SSR offers a consistent and transferable
post-processing signal rather than optimizing a single metric
or dataset.

\paragraph{Comparison with Training-Free Post-Processing.}

DenseCRF, LP, SS-Ncut, and SSR refine the same frozen slot assignments
but rely on different evidence. DenseCRF primarily sharpens masks according
to local RGB and edge consistency. LP diffuses slot evidence through a
static feature graph, while SS-Ncut combines static feature affinity with a
graph-partitioning transition. SSR instead constructs its graph from the change in pairwise
similarity caused by the attention operator. It asks which patches are
drawn closer together during value aggregation rather than which patches
are already similar in a frozen representation. The results show that this
dynamic signal yields the most consistent all-pixel ARI gains and is
particularly effective on COCO, VOC, and YTVIS-2022.

\paragraph{Transfer Across Base Models.}

To examine whether SSR depends on a particular object-centric
architecture, we apply the same refinement rule and hyperparameters to
multiple independently trained image and video models. The improvements on
SLATE, DINOSAUR, SlotDiffusion, SPOT, DIAS, SmoothSA, VideoSAUR,
SlotContrast, and RandSF.Q show that the relation-change signal can be used
with different initial slot assignments without modifying the underlying
model. The DIAS results are particularly useful because DIAS differs from
SmoothSA in its slot initialization and refinement procedure. SSR
improves the DIAS results on VOC, indicating that the observed gains are
not specific to one SmoothSA checkpoint or update rule. We therefore
describe SSR as a transferable post-hoc signal rather than a universal
mask-refinement rule.

\begin{figure*}[t]
\centering
\includegraphics[width=0.9\textwidth]{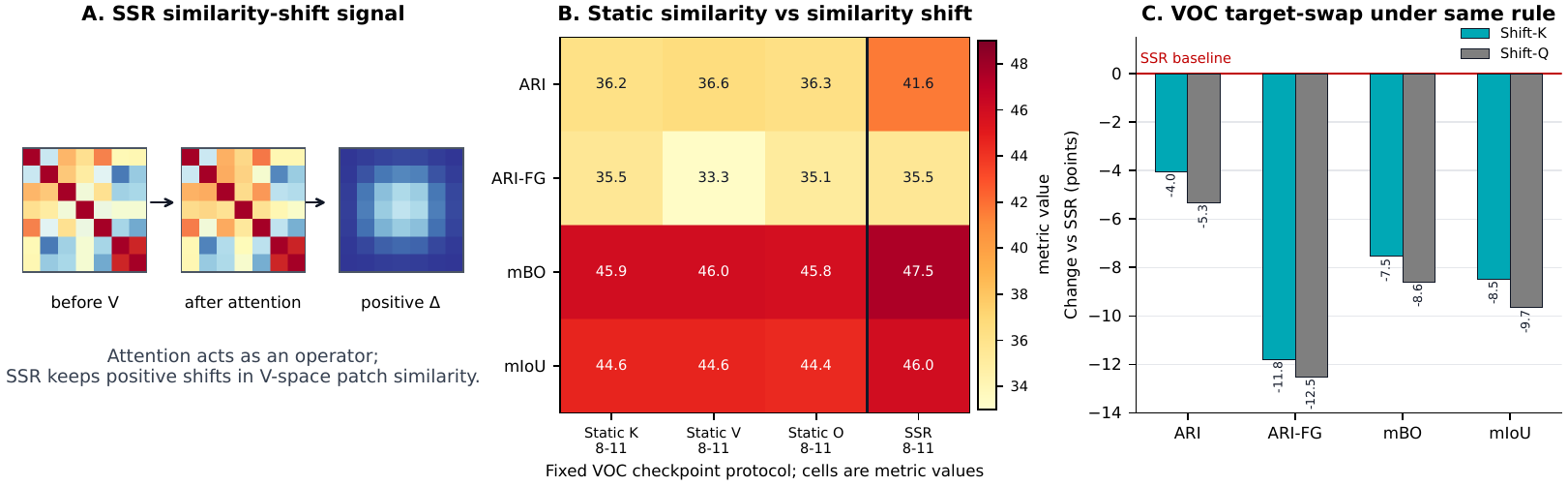}
\caption{
\textbf{Mechanism analysis of SSR.}
(A) SSR retains positive pairwise similarity changes after value
aggregation.
(B) Dynamic $\Delta$-$V$ relations improve ARI and overlap over static
key-, value-, and output-space affinities while remaining competitive on
ARI-FG.
(C) Applying the same construction to queries or keys degrades foreground
and overlap quality.
}
\label{fig:ssr_signal_analysis}
\end{figure*}

\begin{table}[t]
  \centering
\caption{
VOC analysis over three SmoothSA checkpoints (mean$\pm$std); bold marks
the best per panel. SSR uses reliability fusion, patch-only dense attention,
$k=48$, $\tau=0.1$, $\alpha=0.75$, and layers 8--11.
}
  \label{tab:compact_analysis}
  \scriptsize
  \setlength{\tabcolsep}{1.2pt}
  \renewcommand{\arraystretch}{0.94}

  \resizebox{\columnwidth}{!}{%
  \begin{tabular}{@{}lcccrrrr@{}}
  \toprule

  \multicolumn{8}{@{}l}{
    \textit{A. Graph signal, fusion, and attention construction}
  } \\
  \multicolumn{4}{@{}l}{}
  & ARI & ARI-FG & mBO & mIoU \\
  \midrule

  \multicolumn{8}{@{}l}{\itshape Graph-signal controls} \\

  \multicolumn{4}{@{}l}{Aligned masks ($\alpha=0$)}
  & 34.11{\scriptsize$\pm$0.41}
  & 33.53{\scriptsize$\pm$0.88}
  & 44.97{\scriptsize$\pm$0.04}
  & 43.73{\scriptsize$\pm$0.05} \\

  \multicolumn{4}{@{}l}{Post-attention $Z$}
  & 36.16{\scriptsize$\pm$0.33}
  & 35.12{\scriptsize$\pm$1.13}
  & 45.72{\scriptsize$\pm$0.24}
  & 44.34{\scriptsize$\pm$0.29} \\

  \multicolumn{4}{@{}l}{
    Overlap $\mathbf{A}\mathbf{A}^{\mathsf T}$
  }
  & 37.98{\scriptsize$\pm$0.31}
  & 34.55{\scriptsize$\pm$1.63}
  & 46.47{\scriptsize$\pm$0.18}
  & 45.07{\scriptsize$\pm$0.21} \\

  \multicolumn{4}{@{}l}{Shuffled shift}
  & 31.81{\scriptsize$\pm$0.88}
  & 20.99{\scriptsize$\pm$0.28}
  & 33.39{\scriptsize$\pm$0.12}
  & 28.69{\scriptsize$\pm$0.10} \\

  \addlinespace[2pt]
  \multicolumn{8}{@{}l}{\itshape SSR design choices} \\

  \multicolumn{4}{@{}l}{Uniform fusion}
  & 41.74{\scriptsize$\pm$0.35}
  & 35.10{\scriptsize$\pm$1.16}
  & 47.45{\scriptsize$\pm$0.37}
  & 45.93{\scriptsize$\pm$0.43} \\

  \multicolumn{4}{@{}l}{Reliability fusion (default)}
  & 41.62{\scriptsize$\pm$0.31}
  & \textbf{35.52}{\scriptsize$\pm$1.17}
  & 47.46{\scriptsize$\pm$0.34}
  & 45.95{\scriptsize$\pm$0.41} \\

  \multicolumn{4}{@{}l}{All-token aggregation}
  & \textbf{41.95}{\scriptsize$\pm$0.32}
  & 34.74{\scriptsize$\pm$1.07}
  & \textbf{47.53}{\scriptsize$\pm$0.36}
  & \textbf{46.02}{\scriptsize$\pm$0.43} \\

  \multicolumn{4}{@{}l}{Top-64 attention}
  & 40.86{\scriptsize$\pm$0.41}
  & 35.14{\scriptsize$\pm$1.75}
  & 47.14{\scriptsize$\pm$0.36}
  & 45.61{\scriptsize$\pm$0.42} \\

  \midrule
\multicolumn{8}{@{}l}{
  \textit{B. Local sensitivity
  (default: $k=48$, $\tau=0.10$, $\alpha=0.75$)}
} \\
& $k$ & $\tau$ & $\alpha$
& ARI & ARI-FG & mBO & mIoU \\
  \midrule

  & 48 & 0.10 & 0.75
& 41.62{\scriptsize$\pm$0.31}
& 35.52{\scriptsize$\pm$1.17}
& 47.46{\scriptsize$\pm$0.34}
& 45.95{\scriptsize$\pm$0.41} \\

  & 48 & 0.05 & 0.75
  & 41.30{\scriptsize$\pm$0.40}
  & \textbf{35.70}{\scriptsize$\pm$1.20}
  & 47.50{\scriptsize$\pm$0.40}
  & 46.00{\scriptsize$\pm$0.40} \\

  & 56 & 0.10 & 0.75
  & 41.80{\scriptsize$\pm$0.30}
  & 35.30{\scriptsize$\pm$0.90}
  & \textbf{47.60}{\scriptsize$\pm$0.40}
  & \textbf{46.10}{\scriptsize$\pm$0.50} \\

  & 40 & 0.10 & 0.75
  & 41.20{\scriptsize$\pm$0.40}
  & 35.50{\scriptsize$\pm$1.30}
  & 47.40{\scriptsize$\pm$0.40}
  & 45.90{\scriptsize$\pm$0.40} \\

  & 48 & 0.20 & 0.75
  & 41.60{\scriptsize$\pm$0.20}
  & 35.50{\scriptsize$\pm$0.90}
  & 47.50{\scriptsize$\pm$0.40}
  & 46.00{\scriptsize$\pm$0.40} \\

  & 48 & 0.10 & 0.80
  & \textbf{42.10}{\scriptsize$\pm$0.30}
  & 34.60{\scriptsize$\pm$0.80}
  & 47.50{\scriptsize$\pm$0.50}
  & 45.90{\scriptsize$\pm$0.60} \\

  \midrule
  \multicolumn{8}{@{}l}{\textit{C. Layer range}} \\
  \multicolumn{4}{@{}l}{}
  & ARI & ARI-FG & mBO & mIoU \\
  \midrule

  \multicolumn{4}{@{}l}{Layers 8--11 (default)}
& 41.62{\scriptsize$\pm$0.31}
& \textbf{35.52}{\scriptsize$\pm$1.17}
& 47.46{\scriptsize$\pm$0.34}
& 45.95{\scriptsize$\pm$0.41} \\

  \multicolumn{4}{@{}l}{All layers}
  & \textbf{43.30}{\scriptsize$\pm$0.10}
  & 32.70{\scriptsize$\pm$2.00}
  & 47.50{\scriptsize$\pm$0.50}
  & 45.90{\scriptsize$\pm$0.50} \\

  \multicolumn{4}{@{}l}{All layers, cap 2/layer}
  & 42.40{\scriptsize$\pm$0.40}
  & 32.80{\scriptsize$\pm$1.60}
  & 47.90{\scriptsize$\pm$0.40}
  & 46.30{\scriptsize$\pm$0.40} \\

  \multicolumn{4}{@{}l}{All layers, cap 4/layer}
  & 42.60{\scriptsize$\pm$0.40}
  & 33.40{\scriptsize$\pm$2.10}
  & 48.00{\scriptsize$\pm$0.40}
  & 46.50{\scriptsize$\pm$0.40} \\

  \multicolumn{4}{@{}l}{Layers 6--11}
  & 41.60{\scriptsize$\pm$0.50}
  & 35.30{\scriptsize$\pm$1.40}
  & 47.70{\scriptsize$\pm$0.40}
  & 46.20{\scriptsize$\pm$0.50} \\

  \multicolumn{4}{@{}l}{Layers 4--11, cap 4/layer}
  & 41.40{\scriptsize$\pm$0.40}
  & 35.20{\scriptsize$\pm$1.40}
  & \textbf{48.10}{\scriptsize$\pm$0.40}
  & \textbf{46.60}{\scriptsize$\pm$0.50} \\

  \bottomrule
  \end{tabular}%
  }
\end{table}

\section{Ablation Study and Analysis}
\label{sec:analysis}

We analyze SSR from five perspectives: the choice of
value-space relational shift, validity controls against
resizing and generic attention smoothing, attention
construction and layer--head fusion, hyperparameter
robustness, and failure mitigation on ClevrTex. These
experiments clarify the source of SSR's gains, its robustness,
and the conditions under which graph-safety constraints are
needed.



\paragraph{Why Value-Space Relational Change?}

Figure~\ref{fig:ssr_signal_analysis} compares dynamic
value-space shifts with static affinities and alternative
target spaces. Static key-, value-, and output-space graphs
obtain $36.2$--$36.6$ ARI, whereas $\Delta$-$V$ reaches
$41.6$ and also improves mBO and mIoU. This shows that the
gain does not arise from propagating masks over frozen
Transformer similarities alone. Replacing
$\mathbf{A}_{\ell h}\mathbf{V}_{\ell h}$ with
$\mathbf{A}_{\ell h}\mathbf{K}_{\ell h}$ or
$\mathbf{A}_{\ell h}\mathbf{Q}_{\ell h}$ reduces ARI-FG by
$11.8$ and $12.5$ points and mIoU by $8.5$ and $9.7$
points, respectively. Queries and keys primarily determine
attention routing, whereas values carry the content mixed by
attention. These results support value-space relational change
as a more suitable refinement signal than static affinity or
routing-space alternatives.

\paragraph{Validity and Graph-Signal Controls.}

Panel A of Table~\ref{tab:compact_analysis} tests whether
SSR's improvement can instead be attributed to mask
resampling, generic attention smoothing, or shared attention
neighborhoods. The aligned identity baseline follows the same
resize--renormalize--upsample pipeline as SSR but disables
propagation by setting $\alpha=0$. It obtains
$34.11{\pm}0.41$ ARI, compared with
$41.62{\pm}0.31$ for SSR, showing that mask alignment alone
does not explain the improvement. Using post-attention similarity
$\cos(\mathbf{Z}_i,\mathbf{Z}_j)$ without subtracting the
original value-space similarity reaches
$36.16{\pm}0.33$ ARI. Attention-row overlap
$\mathbf{A}\mathbf{A}^{\top}$ performs somewhat better at
$37.98{\pm}0.31$, but both remain substantially below SSR.
Moreover, spatially shuffling the attention correspondence
reduces ARI to $31.81{\pm}0.88$, ARI-FG to
$20.99{\pm}0.28$, mBO to $33.39{\pm}0.12$, and mIoU to
$28.69{\pm}0.10$. These controls indicate that SSR does not merely exploit
preprocessing, post-attention smoothing, or shared attention
patterns. Its effectiveness depends on preserving the spatial
correspondence of the change between pre- and
post-aggregation value relations.

\paragraph{Attention Construction, Sparsification, and
Layer--Head Fusion.}
Panel A of Table~\ref{tab:compact_analysis} examines whether SSR
depends on the treatment of special tokens, the density of the
attention aggregation, or the proposed layer--head weighting
scheme. Including class and register tokens during value
aggregation slightly increases ARI from $41.62$ to $41.95$
and mIoU from $45.95$ to $46.02$, but reduces ARI-FG from
$35.52$ to $34.74$. Thus, prefix tokens mainly shift the
balance toward global clustering rather than explaining the
SSR improvement. We retain patch-only aggregation as the
default because it provides the strongest foreground score
and defines a direct patch-to-patch refinement operator.

Restricting each patch to its Top-$64$ attention neighbors
reduces ARI to $40.86$ and lowers both overlap metrics,
suggesting that sparsifying the attention operator before
computing the relational shift removes useful contextual
information. SSR therefore applies Top-$k$ sparsification to
the resulting similarity-shift graph rather than to the
attention matrix itself. Finally, replacing reliability-weighted layer--head fusion
with a uniform mean produces nearly identical performance:
ARI-FG decreases from $35.52$ to $35.10$, while ARI, mBO, and mIoU change by at most $0.12$ points. This result shows
that the main gain arises from the value-space
similarity-shift signal rather than from a specialized
layer--head weighting rule. Reliability weighting is retained
as the default because it provides a slightly better balance
of foreground and overlap metrics.

\paragraph{Hyperparameter Robustness and Layer Selection.}
Panel B of Table~\ref{tab:compact_analysis} varies one
hyperparameter at a time around the default configuration.
Changing the neighborhood size from $k=48$ to $k=40$
slightly reduces ARI from $41.6$ to $41.2$, whereas
increasing it to $k=56$ raises ARI to $41.8$ but slightly
reduces ARI-FG. This indicates that larger neighborhoods
provide only marginal global-clustering gains while
increasing the risk of cross-instance propagation. The method is insensitive to the layer-head weighting
temperature: $\tau\in\{0.05,0.1,0.2\}$ produces nearly
identical overlap scores and only small ARI variations.
Increasing the propagation strength from $\alpha=0.75$ to
$\alpha=0.80$ raises ARI from $41.6$ to $42.1$ but reduces
ARI-FG from $35.5$ to $34.6$. We therefore select
$k=48$, $\tau=0.1$, and $\alpha=0.75$ as a balanced
operating point. Panel C compares different DINOv2 layer ranges. Using all
layers increases all-pixel ARI but reduces ARI-FG, suggesting
that earlier layers introduce local appearance and texture
relations that do not always respect instance boundaries.
Capped broader ranges slightly improve overlap metrics but
retain the same global-clustering versus foreground-
preservation trade-off. We consequently use layers 8--11 as
the default balance between global grouping, foreground
quality, and implementation simplicity.

\paragraph{Failure Boundary on Texture-Dense Scenes.}

\begin{table}[th]
  \centering
\caption{
ClevrTex graph-safety results over three checkpoints.
Directed SSR uses the original incoming propagation
$\mathbf{D}^{\top}\mathbf{M}$. Mutual-$k$NN retains only
reciprocal edges. Semantic/local-boundary SSR additionally
uses frozen token similarity, adaptive update strength, and
local pseudo-superpixel boundaries.
}
  \label{tab:clevrtex_graph_safety}
  \scriptsize
  \setlength{\tabcolsep}{3.0pt}
  \renewcommand{\arraystretch}{1.08}
  \resizebox{\columnwidth}{!}{%
  \begin{tabular}{@{}llcccc@{}}
  \toprule
  Model & Variant & ARI & ARI-FG & mBO & mIoU \\
  \midrule
  \multirow{4}{*}{SmoothSA}
  & Aligned identity
  & $76.89{\pm}0.74$
  & $\mathbf{82.20{\pm}1.69}$
  & $\mathbf{60.64{\pm}0.44}$
  & $\mathbf{58.90{\pm}0.62}$ \\

  & Directed SSR
  & $\mathbf{82.15{\pm}0.23}$
  & $14.50{\pm}2.44$
  & $25.60{\pm}1.87$
  & $23.84{\pm}1.99$ \\

  & Mutual-$k$NN SSR
  & $77.63{\pm}0.13$
  & $58.59{\pm}3.48$
  & $47.11{\pm}1.84$
  & $44.55{\pm}2.01$ \\

  & Semantic/local-boundary SSR
  & $81.19{\pm}0.27$
  & $73.89{\pm}3.28$
  & $59.51{\pm}1.77$
  & $57.45{\pm}1.97$ \\
  \midrule
  \multirow{4}{*}{SPOT}
  & Aligned identity
  & $25.56{\pm}1.25$
  & $77.10{\pm}0.51$
  & $48.24{\pm}0.50$
  & $46.32{\pm}0.61$ \\
  & Directed SSR
  & $\mathbf{41.86{\pm}6.35}$
  & $56.23{\pm}1.31$
  & $45.26{\pm}1.34$
  & $43.01{\pm}1.20$ \\
  & Mutual-$k$NN SSR
  & $31.61{\pm}1.08$
  & $72.03{\pm}0.58$
  & $46.30{\pm}0.20$
  & $43.86{\pm}0.22$ \\
  & Semantic/local-boundary SSR
  & $29.62{\pm}0.92$
  & $\mathbf{78.11{\pm}0.49}$
  & $\mathbf{50.60{\pm}0.57}$
  & $\mathbf{48.42{\pm}0.67}$ \\
  \bottomrule
  \end{tabular}%
  }
  \end{table}

\begin{figure}[th]
\centering
\includegraphics[width=0.95\columnwidth]{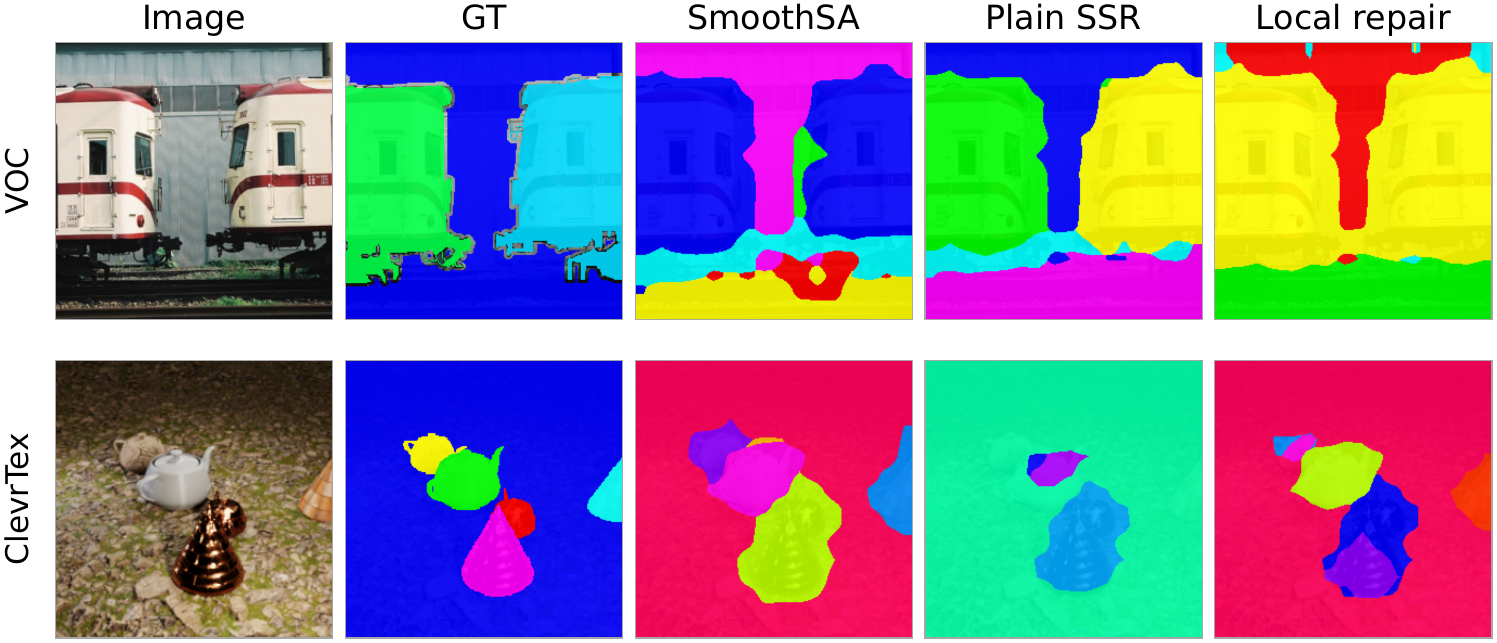}
\caption{
\textbf{Qualitative behavior of SSR on VOC and ClevrTex.}
 On VOC, relation propagation repairs fragmented regions and improves consistency. On ClevrTex, unconstrained propagation may connect texture-similar regions across distinct instances, motivating reciprocal graph sparsification. The semantic/local-boundary variant suppresses such
cross-instance propagation and recovers foreground separation.
}
\label{fig:qualitative_results}
\end{figure}

Table~\ref{tab:clevrtex_graph_safety} and
Figure~\ref{fig:qualitative_results} analyze texture-induced
over-grouping on ClevrTex using two base models, SmoothSA
and SPOT. For both models, directed SSR improves all-pixel
ARI but substantially degrades foreground and overlap
metrics. Mutual-$k$NN partially recovers this loss by
removing nonreciprocal edges.

Semantic/local-boundary SSR provides the strongest recovery.
For SmoothSA, it raises ARI-FG from $14.50$ to $73.89$ and
mIoU from $23.84$ to $57.45$, while retaining an ARI of
$81.19$. For SPOT, it improves all four metrics over the
aligned identity baseline, reaching $29.62$ ARI, $78.11$
ARI-FG, $50.60$ mBO, and $48.42$ mIoU. These results show that reciprocal sparsification and local
semantic-boundary constraints substantially reduce
texture-driven cross-instance propagation. However, the
foreground performance of SmoothSA is not fully restored,
so SSR is best viewed as a global clustering refiner with
optional graph-safety constraints.

\paragraph{Computational Cost.}

For $|\mathcal{L}|$ selected layers, $H$ attention heads,
$N$ patches, and head dimension $d_h$, constructing the
pairwise similarity-shift graphs requires
$O(|\mathcal{L}|HN^2d_h)$ time and
$O(|\mathcal{L}|HN^2)$ temporary storage. SSR introduces no
learned parameters and requires no additional training. As shown in Table~\ref{tab:posthoc_efficiency}, SSR increases
latency from $4.06{\pm}0.03$ to $4.66{\pm}0.03$ ms per
image, corresponding to a $14.8\%$ overhead on the evaluated
setup. Peak allocated CUDA memory increases by $228.9$ MiB.
DenseCRF requires $84.73{\pm}0.28$ ms per image, or
$20.87\times$ the Base latency. These measurements use one
SmoothSA$_i$ seed-42 checkpoint, batch size 16, and a Tesla
V100S.
\begin{table}[t]
\centering
  \caption{Efficiency on VOC validation using the same frozen
  SmoothSA$_i$ checkpoint, batch size 16, and one Tesla V100S.
  Values are mean$\pm$standard deviation over three interleaved repeats;
  CUDA memory denotes peak allocated memory.}
  \label{tab:posthoc_efficiency}

  \resizebox{\columnwidth}{!}{%
    \begin{tabular}{lcccc}
      \toprule
      Method
      & Latency (ms/image) $\downarrow$
      & vs.\ Base
      & Throughput (image/s) $\uparrow$
      & Peak CUDA (MiB) $\downarrow$ \\
      \midrule

      Base
      & $4.06{\pm}0.03$
      & $1.00{\times}$
      & $246.27{\pm}2.03$
      & 489.3 \\

      SSR
      & $4.66{\pm}0.03$
      & $1.15{\times}$
      & $214.70{\pm}1.30$
      & 718.2 \\

      DenseCRF
      & $84.73{\pm}0.28$
      & $20.87{\times}$
      & $11.80{\pm}0.04$
      & 513.6 \\

      \bottomrule
    \end{tabular}%
  }
\end{table}

\section{Conclusion}

We introduce SSR, a training-free post-hoc method that refines
object-centric masks using attention-induced changes in value-space patch
relationships. Across multiple image and video baselines, SSR provides
consistent gains in all-pixel ARI. Matched controls show that the gain is
not explained by mask resizing, post-attention smoothing, attention
overlap, or shuffled spatial correspondence. Texture-dense ClevrTex scenes
reveal that unconstrained propagation may over-group distinct foreground
instances; reciprocal sparsification substantially mitigates this failure
on two base models but does not restore every metric. SSR is therefore
best viewed as a transferable global clustering refiner rather than a
universally safe instance-mask refiner.


\bibliography{aaai2027}


\end{document}